\documentclass[10pt,twocolumn,letterpaper]{article}

\usepackage[pagenumbers]{cvpr} 

\usepackage[dvipsnames]{xcolor}

\definecolor{cvprblue}{rgb}{0.21,0.49,0.74}
\usepackage[pagebackref,breaklinks,colorlinks,citecolor=cvprblue]{hyperref}

\newcommand{\R}{\mathbb{R}}
\newcommand{\bI}{\mathbf{I}}
\newcommand{\by}{\mathbf{y}}
\newcommand{\bys}{\mathbf{y}^*}
\newcommand{\yi}{y_i}
\newcommand{\yis}{y_i^*}
\newcommand{\zi}{z_i}
\newcommand{\di}{d_i}
\newcommand{\bz}{\mathbf{z}}
\newcommand{\bW}{\mathbf{W}}
\newcommand{\bx}{\mathbf{x}}
\newcommand{\nin}{n_{in}}
\newcommand{\nout}{n_{out}}
\newcommand{\N}{\mathcal{N}}
\newcommand{\bB}{\mathbf{B}}

\DeclareMathOperator{\ReLU}{ReLU}
\DeclareMathOperator{\BatchNorm}{BatchNorm}
\DeclareMathOperator{\E}{E}
\DeclareMathOperator{\Var}{Var}

\def\secref#1{Section~\ref{#1}}
\def\figref#1{Figure~\ref{#1}}
\def\tabref#1{Table~\ref{#1}}
\def\eqref#1{Eq.~\ref{#1}}

\usepackage{derivative}
\usepackage{lipsum}
\usepackage{graphicx}

\usepackage{listings}
\usepackage{xcolor}
\usepackage{multirow}
\usepackage{mathtools}  

\definecolor{codegreen}{rgb}{0,0.6,0}
\definecolor{codegray}{rgb}{0.5,0.5,0.5}
\definecolor{codepurple}{rgb}{0.58,0,0.82}
\definecolor{backcolour}{rgb}{0.95,0.95,0.92}

\lstdefinestyle{mystyle}{
    backgroundcolor=\color{backcolour},
    commentstyle=\color{codegreen},
    keywordstyle=\color{magenta},
    numberstyle=\tiny\color{codegray},
    stringstyle=\color{codepurple},
    basicstyle=\ttfamily\footnotesize,
    breakatwhitespace=false,
    breaklines=true,
    captionpos=b,
    keepspaces=true,
    numbers=left,
    numbersep=5pt,
    showspaces=false,
    showstringspaces=false,
    showtabs=false,
    tabsize=2
}

\def\paperID{*****} 
\def\confName{CVPR}
\def\confYear{2024}

\title{Analysis of NaN Divergence in Training Monocular Depth Estimation Model}

\author{Bum Jun Kim\\
POSTECH\\
{\tt\small kmbmjn@postech.ac.kr}
\and
Hyeonah Jang\\
POSTECH\\
{\tt\small hajang@postech.ac.kr}
\and
Sang Woo Kim\\
POSTECH\\
{\tt\small swkim@postech.ac.kr}
}

\begin{document}
\maketitle
\begin{abstract}
	The latest advances in deep learning have facilitated the development of highly accurate monocular depth estimation models. However, when training a monocular depth estimation network, practitioners and researchers have observed not a number (NaN) loss, which disrupts gradient descent optimization. Although several practitioners have reported the stochastic and mysterious occurrence of NaN loss that bothers training, its root cause is not discussed in the literature. This study conducted an in-depth analysis of NaN loss during training a monocular depth estimation network and identified three types of vulnerabilities that cause NaN loss: 1) the use of square root loss, which leads to an unstable gradient; 2) the log-sigmoid function, which exhibits numerical stability issues; and 3) certain variance implementations, which yield incorrect computations. Furthermore, for each vulnerability, the occurrence of NaN loss was demonstrated and practical guidelines to prevent NaN loss were presented. Experiments showed that both optimization stability and performance on monocular depth estimation could be improved by following our guidelines.
\end{abstract}

\section{Introduction}
\label{sec:intro}

Understanding a driving scene is an important task for self-driving or advanced driver-assistance systems. In particular, constructing 3D scene information in a driving environment is crucial. To this end, studying monocular depth estimation, which aims to obtain the physical depth information from a single RGB image, is of paramount importance. Recently, deep neural networks \citep{DBLP:conf/iclr/DosovitskiyB0WZ21,DBLP:conf/cvpr/HeZRS16,DBLP:conf/eccv/ChenZPSA18,DBLP:conf/iccv/RanftlBK21} have been extensively used, and they have also been incorporated into the monocular depth estimation task, in which a deep neural network is trained by the iterative optimization of gradient descent using a large dataset of RGB-LiDAR image pairs. The latest advances in deep learning have allowed us to develop a highly accurate monocular depth estimation model that is close to real-world applications.

However, the training of monocular depth estimation models has stability issues. When training a monocular depth estimation model, practitioners and researchers have observed \texttt{not a number} (NaN) loss, which indicates a failure mode where the optimization diverged owing to inappropriate computations, such as division by zero. Several practitioners have reported NaN loss and difficulties in re-implementing monocular depth estimation models in the official GitHub repositories. When training a monocular depth estimation model, the optimization can either finish normally or lead to a NaN loss, even with the same model and hyperparameter setup. NaN losses have been reported to occur during the early training phase, as well as the late training phase near the optimal point. The stochastic and mysterious occurrence of the NaN loss has been a cause of concern among practitioners and researchers because it wastes time and resources.

It should be noted that NaN loss occurs rarely for other training tasks such as semantic segmentation, while it occurs frequently for the training task of monocular depth estimation. The root cause of the NaN loss has not been discussed in the literature and remains unclear. Determining the root cause is challenging as it requires debugging the training procedure of the deep neural network. Thus, currently, further research is required to reveal the root cause of NaN loss in training monocular depth estimation models. This is expected to overcome the instability of optimization, and thereby facilitate the research community and practitioners.

In this paper, we report an in-depth analysis that focuses on the occurrence of NaN loss when training a monocular depth estimation model. We thoroughly inspected the current implementations of monocular depth estimation networks, and discovered three types of vulnerabilities in their gradient descent optimization:
\begin{itemize}
	\item First, the use of the square root loss, which is a common practice, yields an exploding gradient when approaching an optimal point. We discuss square root loss and its potential advantages and disadvantages.
	\item Second, the log-sigmoid function used in the monocular depth estimation network has numerical stability issues and is prone to NaN loss. This problem can be addressed either through careful weight initialization or by improving the numerical stability of the logarithmic function. To this end, we present a stable initialization range that assures the absence of NaN loss.
	\item Finally, a critical error exists in the implementation of variance computation, where half of the current implementations of monocular depth estimation models have potential problems. In a practical scenario, we demonstrated the accidental occurrence of NaN loss caused by this vulnerability.
\end{itemize}
For each vulnerability, we analyzed detailed computations and empirically demonstrated the occurrence of NaN through several simulations. We also provide practical guidelines for solving the NaN loss for each vulnerability.

\section{Vulnerability Report}
\label{sec:report}

\begin{figure*}[t!]
	\centering
	\begin{subfigure}{0.33\linewidth}
		\includegraphics[width=0.99\linewidth]{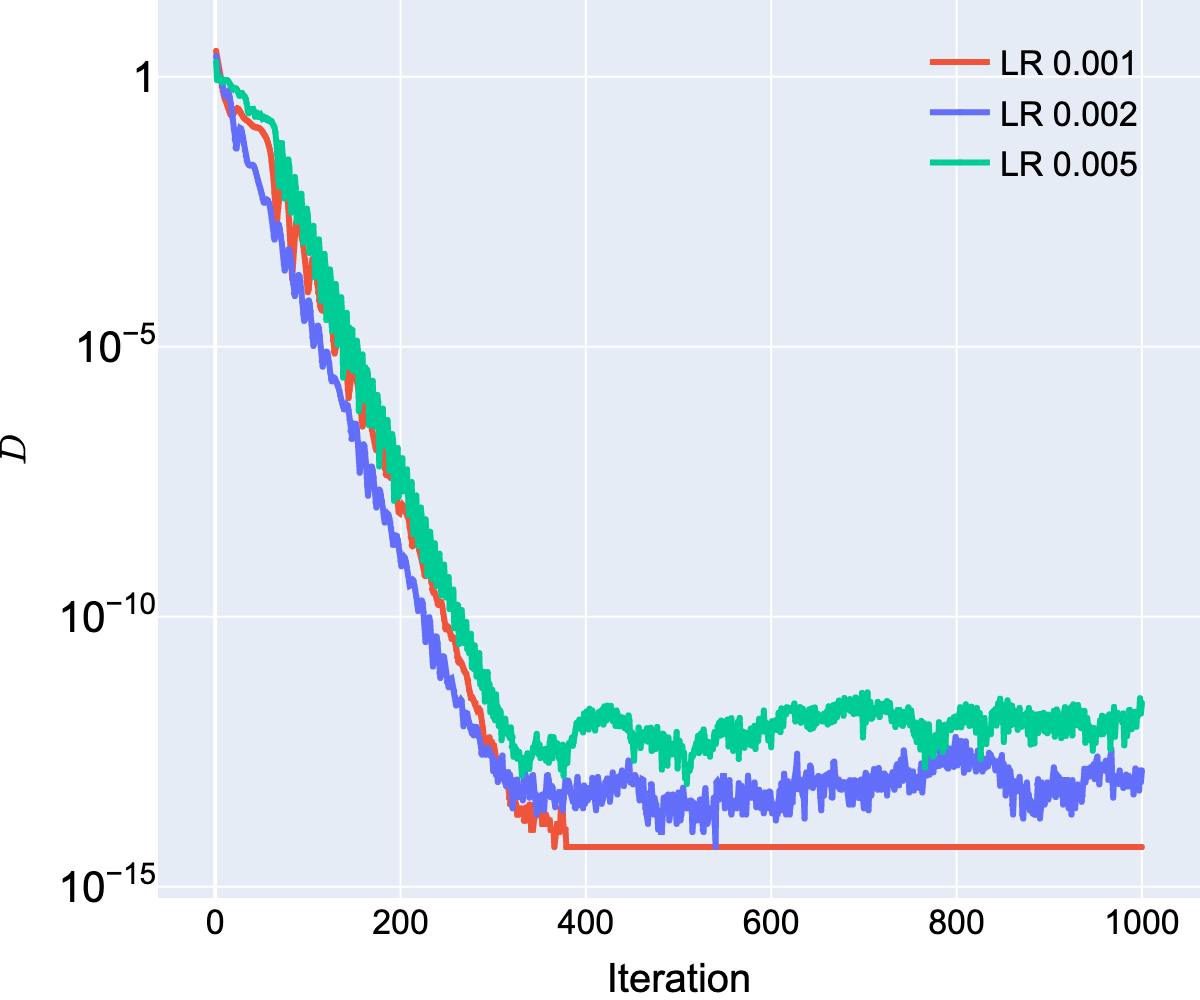}
		\caption{Minimizing $D$.}
		\label{fig:sqrt_a}
	\end{subfigure}
	\hfill
	\begin{subfigure}{0.33\linewidth}
		\includegraphics[width=0.99\linewidth]{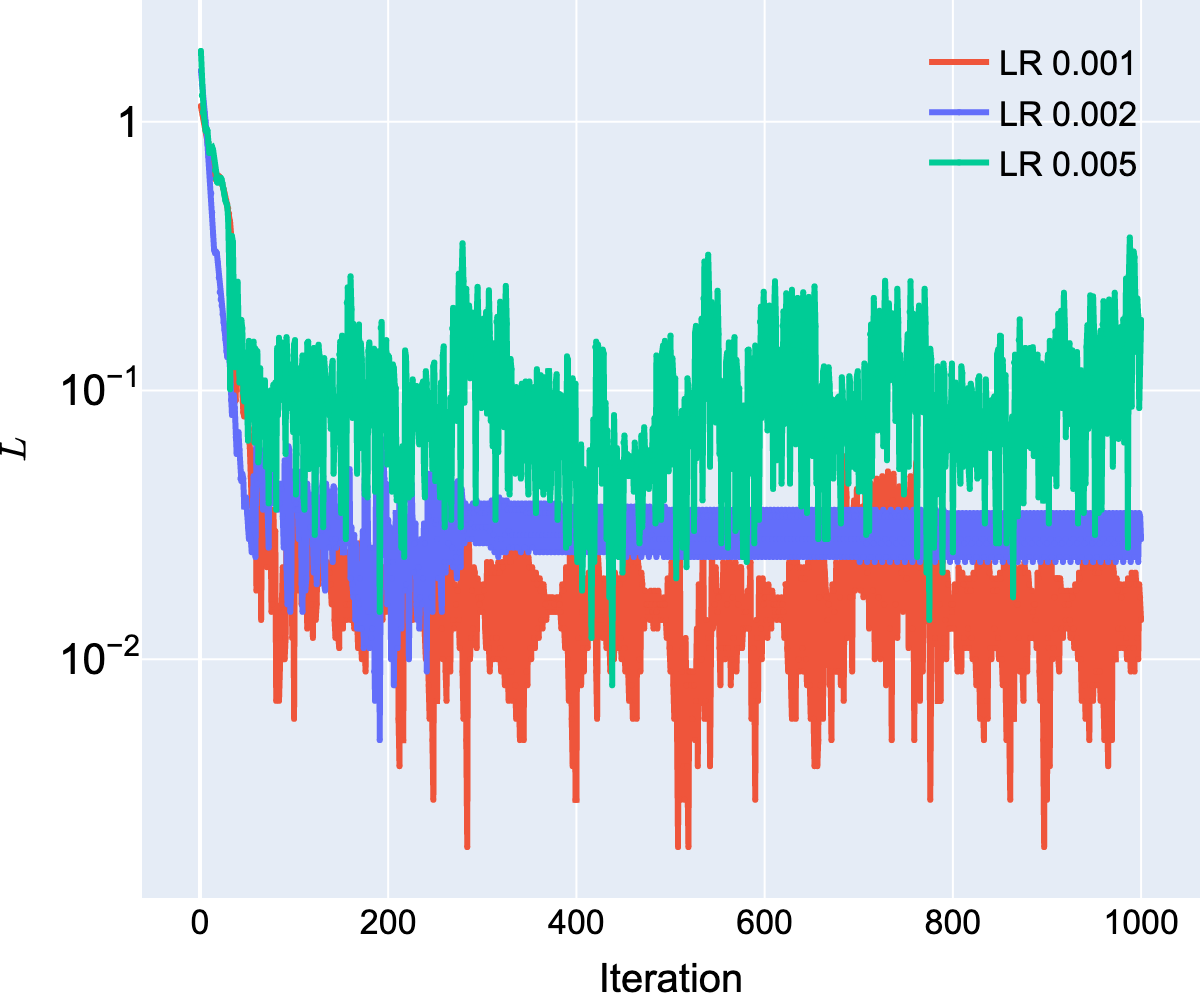}
		\caption{Minimizing $L=\sqrt{D}$.}
		\label{fig:sqrt_b}
	\end{subfigure}
	\hfill
	\begin{subfigure}{0.33\linewidth}
		\includegraphics[width=0.99\linewidth]{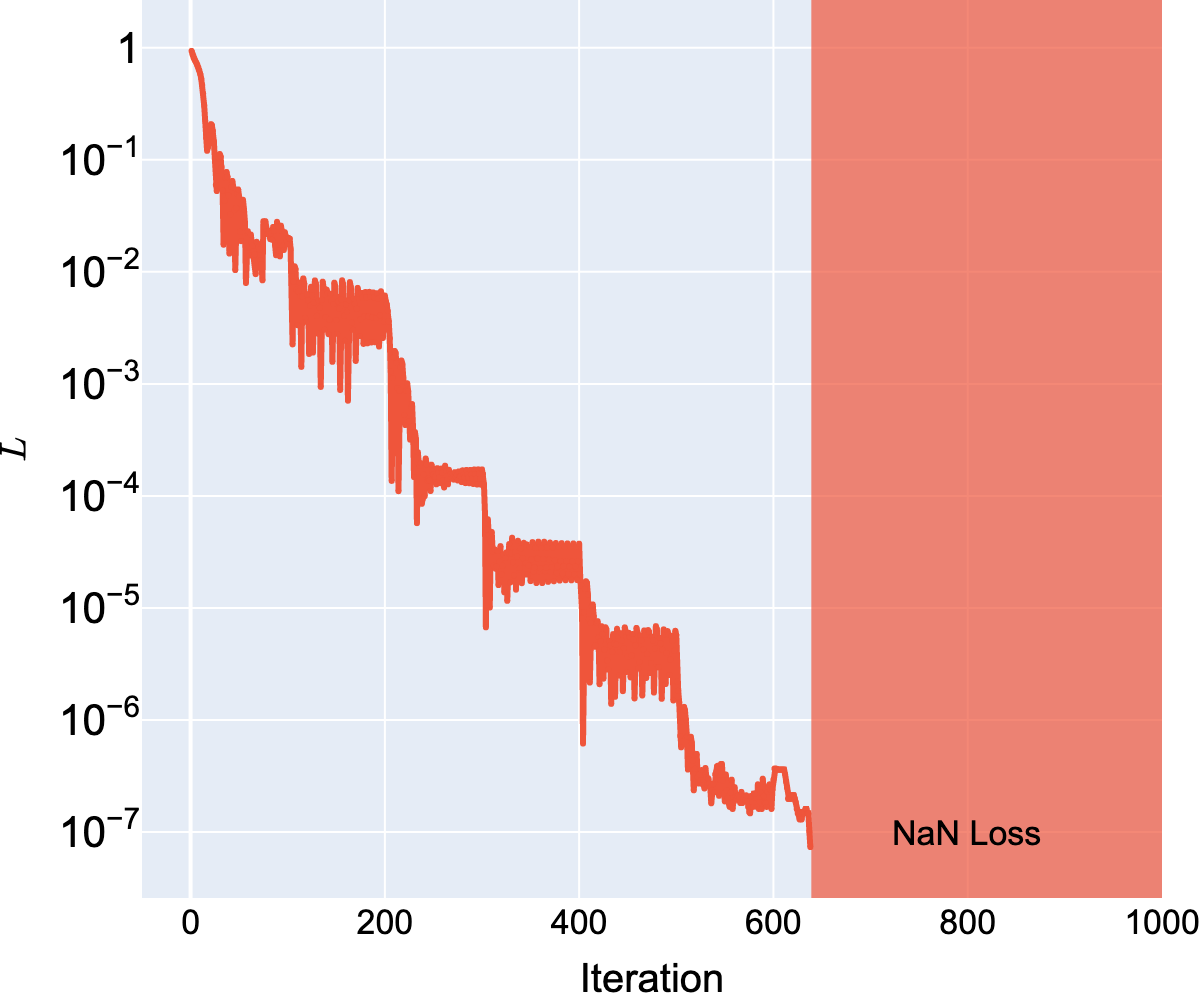}
		\caption{Minimizing $L=\sqrt{D}$ with learning rate decay.}
		\label{fig:sqrt_c}
	\end{subfigure}
	\caption{Simulation results comparing the choice of loss.}
	\label{fig:sqrt}
\end{figure*}

\paragraph{Background and Formulation} This study considers the standard framework for supervised learning of the monocular depth estimation model. Let $\bI \in \R^{n_h \times n_w \times n_c}$ be an input image for a monocular depth estimation model, where $(n_h \times n_w)$ is the size of the image and $n_c$ represents the number of channels. The objective of monocular depth estimation is to generate a depth map $\by \in \R^{n_h \times n_w}$ that estimates physical distance $\yi$ for the $i$-th pixel in the image and exhibits a small error with its ground-truth $\bys$. A deep neural network composed of an encoder and decoder with a head is used as the monocular depth estimation model that outputs $\by$ from the input image $\bI$. The monocular depth estimation task has constraints on the maximum depth $M$ in meters, such as $0 < \yi < 80$, with $M=80$. To generate a depth map with a constrained depth range, a sigmoid head with a max scaler has been commonly used as $\yi = M p_i = M s(\zi)$, where $p = s(z)=1/(1+e^{-z})$. Here, $\bz \in \R^{n_h \times n_w}$ is obtained using the last convolutional layer of the decoder $\bz = \bW\bx + b$, where $\bx \in \R^{n_h \times n_w \times \nin}$, $\bW \in \R^{k_h \times k_w \times \nin \times 1}$, and $b \in \R$.

Monocular depth estimation is a regression task aimed at obtaining real value predictions. However, opposed to common regression tasks, describing relative farness and nearness is more important in this task. Considering this task property, \citet{DBLP:conf/nips/EigenPF14} proposed scale-invariant log loss $D$ as follows:
\begin{align}
	D = \frac{1}{n} \sum_i{d_i^2} - \frac{\lambda}{n^2} (\sum_i{d_i)^2},
\end{align}
where $\di = \log{\yi} - \log{\yis}$, $\lambda \in [0, 1]$, and $n$ denotes the number of valid pixels. The scale-invariant log loss and its gradient do not depend on the maximum depth $M$ and are suitable for reflecting relative farness and nearness. The monocular depth estimation network is optimized by minimizing the scale-invariant log loss, ideally to zero.

In practice, \citet{DBLP:journals/corr/abs-1907-10326} proposed alternatively minimizing the square root of the scale-invariant log loss as follows:
\begin{align}
	L = \sqrt{D}.
\end{align}
They empirically observed that the use of the square root loss led to better performance. Since its introduction, square root loss has been widely employed in monocular depth estimation networks \citep{9824488,DBLP:journals/corr/abs-2302-08149,DBLP:conf/cvpr/PatilSLG22,DBLP:journals/corr/abs-2309-14137,DBLP:conf/iclr/LiuKGTG23}.

\subsection{Vulnerability in Square Root Loss} However, we found that the square root loss caused NaN divergence. This problem arises from the unstable gradient of the square root loss. For weight $\bW$, the gradient descent minimizing square root loss $L$ with learning rate $\eta$ can be expressed as follows:
\begin{align}
	\bW          & \leftarrow \bW - \eta \pdv{L}{\bW},                           \\
	\pdv{L}{\bW} & = \pdv{L}{D} \pdv{D}{\bW} = \frac{1}{2\sqrt{D}} \pdv{D}{\bW}.
\end{align}
By contrast, when using the scale-invariant log loss $D$ for the minimization objective of the gradient descent, the gradient is $\partial D / \partial \bW$. Thus, compared with scale-invariant log loss $D$, the use of square root loss $L$ causes scaling of the gradient by $1/2\sqrt{D}$.

\paragraph{Pros} The use of square root loss may help with optimization during the early training phase. In the early training phase, if the optimization is at an unstable point with a large loss and gradient, the use of square loss leads to a smaller gradient by a large $D$. This behavior reduces the large weight updates, similar to gradient clipping \citep{DBLP:conf/icml/PascanuMB13,DBLP:conf/iclr/ZhangHSJ20}, thereby preventing unstable optimization. Furthermore, to avoid overfitting to the training set, preventing zero loss and maintaining suboptimal loss during training is crucial since training and test losses differ in a strict sense \citep{DBLP:conf/nips/Li0TSG18,DBLP:conf/nips/GaripovIPVW18}. Considering this property, the use of square root loss may be beneficial to avoid overfitting to the training set and produce a type of regularization effect.

\paragraph{Cons} The serious vulnerability of the square root loss arises during the late training phase, when the loss is suitably minimized. Note that the objective of the monocular depth estimation task is to achieve $\by \rightarrow \bys$ and $D \rightarrow 0$ as much as possible. However, as $D$ approaches zero, owing to the presence of $\sqrt{D}$ in the denominator, the gradient becomes larger, which causes a deviation from the optimal point and disrupts the optimization. Even if $D=0$ is achieved by any methods, it causes NaN loss.

\paragraph{Simulation} We empirically demonstrate the vulnerability of square root loss. We generated artificial random normal data, whose ground-truth followed the mean and standard deviation of the KITTI dataset \citep{DBLP:journals/ijrr/GeigerLSU13} statistics. We simulated the training of a sigmoid head that received $[\BatchNorm\--\ReLU]$ output as $\bx = \ReLU(\bB)$ with $B_i \sim \N(0, 1)$ with its size $n_h=n_w=3$. We configured the last convolutional layer such that it had the following properties: number of channels $\nin=128$, kernel size $k_h=k_w=3$, and weight initialization $W \sim \N(0, 0.1^2)$. We used a mini-batch size of 10 and the Adam optimizer \citep{DBLP:journals/corr/KingmaB14} with learning rates of $\{0.001, 0.002, 0.005\}$ and 1000 iterations.

First, when using the scale-invariant log loss $D$, the loss was sufficiently minimized to less than $10^{-10}$ (\figref{fig:sqrt_a}). However, when using the square root loss $L$, the optimization did not proceed to a loss below 0.001 (\figref{fig:sqrt_b}), indicating that employing square root loss hinders optimization.

One may attempt to achieve better optimization by using a learning rate scheduler to reduce the weight updates from a large gradient. Here, we employed a stepwise learning rate scheduler that controls the learning rate starting at 0.001 and decays by 0.1 for every 100 iterations. As shown in \figref{fig:sqrt_c}, decaying the learning rate helped the optimization against a large gradient and yielded a smaller loss during the early training phase. However, during the late training phase with losses less than $10^{-7}$, the NaN loss occurred. This observation is consistent with our analysis that approaching the optimal point causes divergence when using the square root loss.

In summary, the use of the square root loss not only hinders the optimization owing to the large gradient but also causes NaN divergence as the loss approaches zero. In consideration of this observation, we claim the following.
\paragraph{Guideline 1.} Although the use of the square root loss is beneficial during the early training phase, it causes unstable gradient behavior during the late training phase. If NaN loss is observed during the late training phase, consider not using the square root loss or switching to the original scale-invariant log loss.

\subsection{Vulnerability in Log-Sigmoid Function} Another vulnerability that causes NaN divergence is the use of a logarithmic function with a sigmoid head. Here, the output distribution of the sigmoid head is significantly affected by the weight initialization of the last convolutional layer. A larger weight scale $\Var[\bW]$ of the last convolutional layer induces a larger scale in $\bz$, which causes the output of the sigmoid head $\yi = M s(\zi)$ to be polarized to 0 or $M$. Note that the resulting value is used for computing $\di = \log{\yi} - \log{\yis}$, whose logarithmic function is numerically unstable for $\yi=0$ since the logarithm of zero is negative infinity. For example, on the FP32 precision of PyTorch \citep{DBLP:conf/nips/PaszkeGMLBCKLGA19}, the sigmoid function outputs $s(-88)=6.0546 \times 10^{-39}$ and $s(-89)=0$. The logarithmic function yields a NaN value for the latter, which leads to the NaN loss. We observed that this scenario occurs in real situations when training a monocular depth estimation network, particularly during the early training phase. To address this problem, we consider two approaches.

\subsubsection{Approach 1. Weight Initialization} Note that $z_i$ is obtained by the last convolutional layer of the decoder and is calculated using $\bz = \bW\bx + b$. First, it is favorable to achieve a stable scale on the decoder feature $\Var[\bx]$ using an additional batch normalization layer \citep{DBLP:conf/icml/IoffeS15} prior to the last $\ReLU$. Although common practices omit the batch normalization layer when computing $\bx$, we observed that incorporating the batch normalization layer marginally improves the performance and stability. Second, in the case of $b$, we observed that the initialization of $b$ rarely influenced the performance and stability, hence, we used zero initialization following common practice. Finally, because a larger weight scale $\Var[\bW]$ would yield a value such as $z_i =-89$, we should set a smaller weight scale.

Note that an initialization of the weight that is too small hinders gradient descent optimization \citep{DBLP:conf/nips/LewkowyczG20,DBLP:journals/corr/abs-2011-11152,DBLP:journals/corr/Laarhoven17b}. Therefore, we need to quantify an initialization range that ensures no NaN loss from a small scale on $\bz$ as well as stable behavior in the gradient descent. Here, we simulated the behavior of a sigmoid head by controlling a weight scale to measure its impact on the gradient scale.

\paragraph{Simulation} We examine whether the gradient scale $\Var[\partial L / \partial \bW]$ would increase or decrease when the weight scale $\Var[\bW]$ increases. Additionally, we monitor the occurrence of NaN loss. We generated artificial random normal data whose ground-truth followed the mean and standard deviation of the \{KITTI \citep{DBLP:journals/ijrr/GeigerLSU13}, NYU-Depth V2 \citep{DBLP:conf/eccv/SilbermanHKF12}, Driving Stereo \citep{DBLP:conf/cvpr/YangSHDSZ19}, Argoverse \citep{DBLP:conf/cvpr/ChangLSSBHW0LRH19}, DDAD \citep{DBLP:conf/cvpr/GuiziliniAPRG20}\} dataset statistics. We simulated the sigmoid head receiving the $[\BatchNorm\--\ReLU]$ output as $\bx = \ReLU(\bB)$ with $B_i \sim \N(0, 1)$ and $n_h=n_w=100$. The last convolutional layer was configured with $\nin=128$ and $k_h=k_w=3$. A mini-batch size of 10 was used. We measure the gradient scale $\Var[\partial L / \partial \bW]$ by controlling the weight scale $\Var[\bW]=\sigma_W^2$. Considering randomness, we measure the average of 1,000 simulations.

\begin{figure}[t!]
	\centering
	\includegraphics[width=0.99\linewidth]{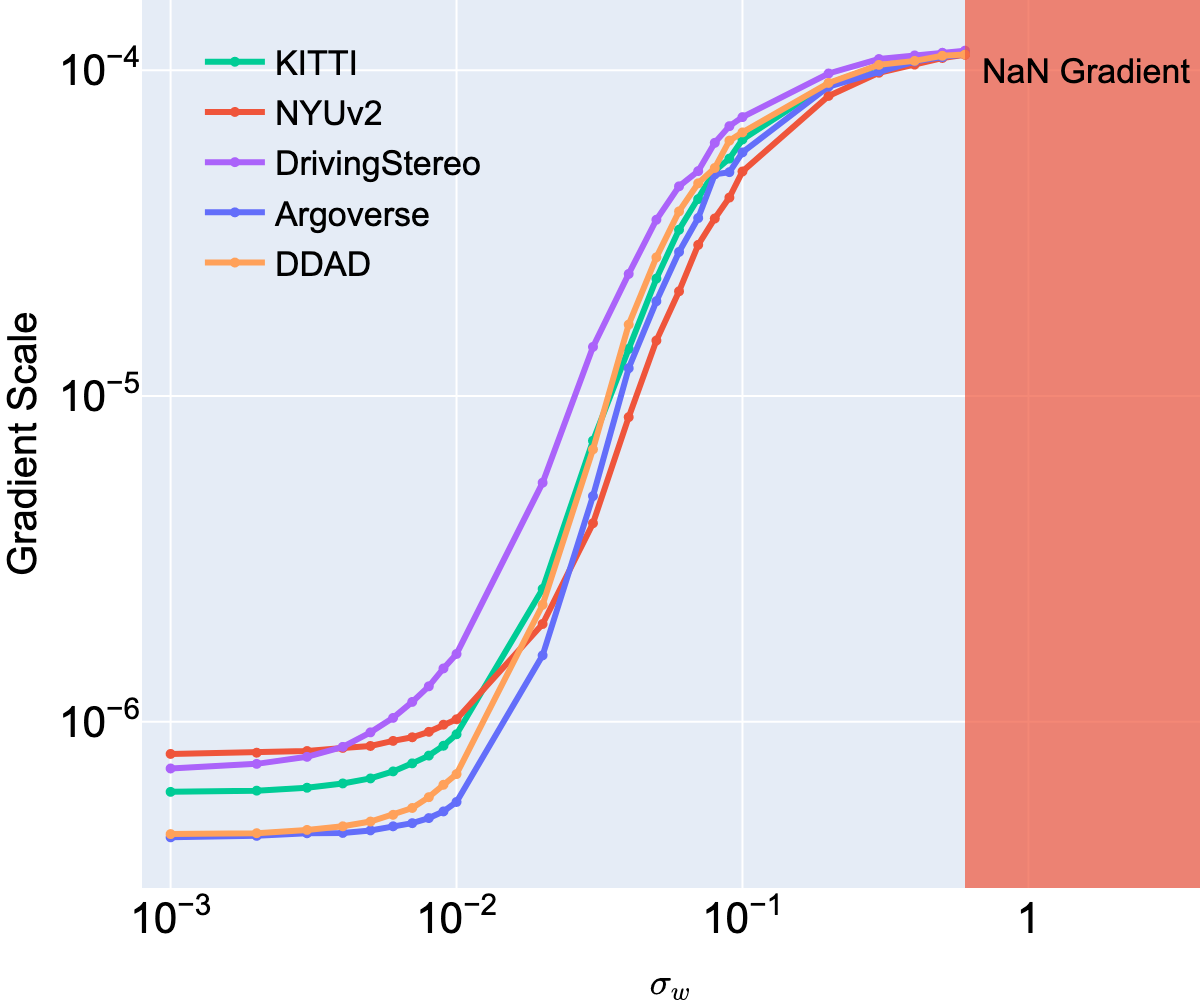}
	\caption{Gradient scale with respect to weight initialization of the last convolutional layer. It should be carefully initialized to achieve a stable gradient scale.}
	\label{fig:gradient_scale}
\end{figure}

\figref{fig:gradient_scale} summarizes the results. Gradient variance is observed to significantly diminish when $\sigma_W < 0.1$, which slows down training similar to the vanishing gradient \citep{DBLP:conf/icml/PascanuMB13}. It should be noted that using the Xavier initialization of $W \sim \N(0, 2/(k_h k_w (\nin + \nout)))$ \citep{DBLP:journals/jmlr/GlorotB10} yields $\sigma_W=0.0415$, which corresponds to this range and therefore should not be used. In contrast, when $\sigma_W > 0.6$, the gradient variance yields a NaN value, which is a result of NaN loss. Stable initialization requires $0.1 \leq \sigma_W \leq 0.6$, which we refer to as \textit{stable initialization range}. This range can be applied to the aforementioned datasets with one exception: the NYU-Depth V2 dataset requires a slightly higher range of $0.2 \leq \sigma_W \leq 0.6$. In \secref{sec:exp}, we demonstrate that weight initialization significantly affects the stability of the optimization as well as the final performance. In summary, the following guidelines are presented:
\paragraph{Guideline 2-1.} The weight of the last convolutional layer should be carefully initialized. Avoid using common weight initialization, such as Xavier initialization. Ensure the use of a weight initialization within the stable initialization range of $0.1 \leq \sigma_W \leq 0.6$.

However, this approach has several limitations. Even though $\sigma_W \leq 0.6$ is applied at initialization, the weight scale can increase during training, which causes NaN loss. Furthermore, since the active gradient scale is close to this boundary of 0.6, we should consider a trade-off between increasing $\sigma_W$ to obtain an active gradient and avoiding potential NaN loss.

\subsubsection{Approach 2. Logarithmic Function}

\begin{figure}[t!]
	\centering
	\includegraphics[width=0.99\linewidth]{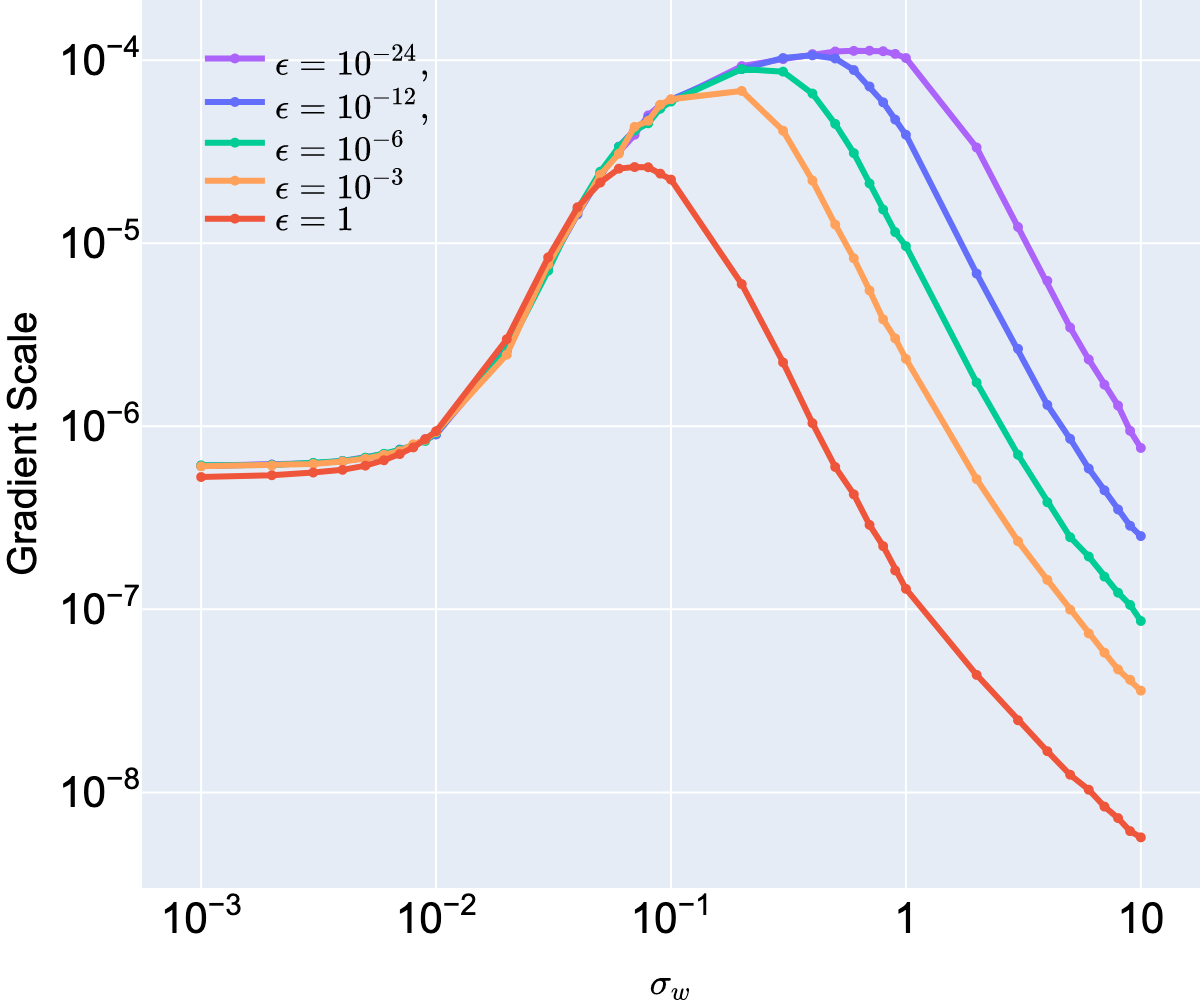}
	\caption{Gradient scale with various choices of $\epsilon$. Adding $\epsilon$ solves the numerical vulnerability of the logarithmic function, unless it is not significantly small such as $\epsilon \leq 7.0\times 10^{-46}$. However, choosing a larger $\epsilon$ decreases the gradient scale.}
	\label{fig:eps}
\end{figure}

Alternatively, we address this vulnerability by improving the numerical stability of the logarithmic function. When using the logarithmic function, adding a small $\epsilon>0$ as $d_{i, \epsilon} \coloneqq \log{(\yi + \epsilon)} - \log{(\yis + \epsilon)}$ ensures that $\yi + \epsilon > 0$ and solves the numerical problem of the logarithmic function. This practice has been employed in a few implementations of monocular depth estimation models with $\epsilon=10^{-3}$ or $\epsilon=10^{-6}$ \citep{DBLP:conf/cvpr/PiccinelliSY23,DBLP:journals/corr/abs-2304-07193,DBLP:journals/corr/abs-2303-17559,DBLP:journals/corr/abs-2203-14211}, while its importance and validity require more emphasis. Subsequently, we investigated the validity of this practice.

\paragraph{Simulation} We used the same simulation setup described above, using the mean and standard deviation of KITTI statistics, but with the addition of $\epsilon$. \figref{fig:eps} summarizes the gradient variance when using $\epsilon \in \{1, 10^{-3}, 10^{-6}, 10^{-12}, 10^{-24}\}$. Furthermore, we observed that $\epsilon \leq 7.0 \times 10^{-46}$ yielded NaN loss because it was smaller than the representable float (See the Appendix for further details), whereas the five values of $\epsilon$ resulted in no instances of NaN.

However, choosing a larger $\epsilon$ decreased the gradient scale. In other words, adding $\epsilon$ approximates the result as $d_{i, \epsilon} \approx d_i$, which affects both the gradient scale and the stable initialization range discussed earlier. In \secref{sec:exp}, we empirically observe that allowing a high gradient scale enhances performance. In consideration of this observation, instead of using $\epsilon=10^{-3}$ or $\epsilon=10^{-6}$, as is the current practice, we recommend using a substantially smaller value such as $\epsilon=10^{-24}$. In summary, an active gradient scale requires a substantially smaller $\epsilon$ with a different stable initialization range. Specifically, we propose the following:
\paragraph{Guideline 2-2.} Adding $\epsilon$ solves the numerical issue of the logarithmic function, but the choice of $\epsilon$ significantly affects the stable initialization range. We recommend using $\epsilon=10^{-24}$ and $0.1 \leq \sigma_W \leq 1$.

\subsection{Vulnerability in Variance Computation}

Rewriting scale-invariant log loss $D$, we obtain
\begin{align}
	D & = \frac{1}{n} \sum_i{d_i^2} - \frac{\lambda}{n^2} (\sum_i{d_i)^2}, \\
	  & = \E[d^2] - \lambda (\E[d])^2 \label{eq:mean}                      \\
	  & = \Var[d] + (1-\lambda) (\E[d])^2. \label{eq:var}
\end{align}
We refer to \eqref{eq:mean} as a mean-style implementation and \eqref{eq:var} as a var-style implementation. The two are identical as long as the variance is calculated as the sum of the squared deviations from the mean divided by $n$, expressed as:
\begin{align}
	\Var[d] = \frac{1}{n} \sum_i{(d_i - \E[d])^2}.
\end{align}

However, for variance computation, modern deep-learning libraries such as PyTorch apply the Bessel correction \citep{reichmann1961use} by default. In other words, \texttt{torch.var(x)} is defined as the sum of the squared deviations from the mean divided by $n-1$, expressed as
\begin{align}
	\Var^{\prime}[d] = \frac{1}{n-1} \sum_i{(d_i - \E[d])^2},
\end{align}
which is also referred to as an unbiased estimator \citep{brown1947small}. Thus, when \texttt{torch.var(x)} is used, the var-style implementation differs from the mean-style implementation. For example, consider $\mathbf{d} = \{1, 3\}$ with $n=2$ and $\lambda=0.5$. This example yields $\Var[d]=1$ and $\Var^{\prime}[d]=2$, resulting in $D=3$ for a biased estimator and $D=4$ for an unbiased estimator. To correctly compute the scale-invariant log loss, we should explicitly specify a biased estimator as \texttt{torch.var(x, unbiased=False)}.

\begin{figure}[t!]
	\begin{lstlisting}[language=Python, caption=Example of mean-style implementation \citep{DBLP:conf/cvpr/XieGH00023}.]
# Reference: https://github.com/SwinTransformer/MIM-Depth-Estimation
class SiLogLoss(nn.Module):
    # ...
    def forward(self, pred, target):
        valid_mask = (target > 0).detach()
        diff_log = torch.log(target[valid_mask]) - torch.log(pred[valid_mask])
        loss = torch.sqrt(torch.pow(diff_log, 2).mean() - self.lambd * torch.pow(diff_log.mean(), 2))
        return loss
\end{lstlisting}
\end{figure}

\begin{figure}[t!]
	\begin{lstlisting}[language=Python, caption=Example of var-style implementation \citep{DBLP:journals/corr/abs-2203-14211}.]
# Reference: https://github.com/zhyever/Monocular-Depth-Estimation-Toolbox
class SigLoss(nn.Module):
    # ...
    def sigloss(self, input, target):
        # ...
        g = torch.log(input + self.eps) - torch.log(target + self.eps)
        Dg = torch.var(g) + 0.15 * torch.pow(torch.mean(g), 2)
        return torch.sqrt(Dg)
\end{lstlisting}
\end{figure}

\begin{table}[t!]
	\centering
	\begin{tabular}{l|l}
		\toprule
		\textbf{Mean-Style}                                & \textbf{Var-Style}                                    \\
		\midrule
		MIM \citep{DBLP:conf/cvpr/XieGH00023}              & DepthFormer \citep{DBLP:journals/corr/abs-2203-14211} \\
		LapDepth \citep{DBLP:journals/tcsv/SongLK21}       & iDisc \citep{DBLP:conf/cvpr/PiccinelliSY23}           \\
		GLPDepth \citep{DBLP:journals/corr/abs-2201-07436} & DDP \citep{DBLP:journals/corr/abs-2303-17559}         \\
		D-Net \citep{DBLP:journals/access/ThompsonPB21}    & Depthformer \citep{DBLP:conf/icip/Agarwal022}         \\
		NeW CRFs \citep{DBLP:conf/cvpr/Yuan0DZT22}         & AdaBins \citep{DBLP:conf/cvpr/BhatAW21}               \\
		PixelFormer \citep{DBLP:conf/wacv/AgarwalA23}      & DINOv2 \citep{DBLP:journals/corr/abs-2304-07193}      \\
		VPD \citep{DBLP:journals/corr/abs-2303-02153}      & ZeoDepth \citep{DBLP:journals/corr/abs-2302-12288}    \\
		AiT \citep{DBLP:journals/corr/abs-2301-02229}      & LocalBins \citep{DBLP:conf/eccv/BhatAW22}             \\
		\bottomrule
	\end{tabular}
	\caption{Half of the current implementations of monocular depth estimation networks used the mean-style implementation, whereas the other half used the var-style implementation. None of the var-style implementations specifies a biased estimator, which results in incorrect loss computation.}
	\label{tab:ref}
\end{table}

We examined the current implementations of monocular depth estimation networks. Half of them used the mean-style implementation, whereas the other half used the var-style implementation without specifying a biased estimator, which resulted in an incorrect loss computation (\tabref{tab:ref}).

When $n \gg 1$, the difference between divisions by $n$ and $n-1$ is insignificant. Vulnerability occurs when $n$ is small, which yields an incorrect loss and gradient. Especially when $n=1$, division by $n-1$ causes NaN loss.

Here, $n$ corresponds to the number of valid pixels during the training. First, valid pixels are dependent on the sparsity of the ground-truth depth map. For example, the ground-truth $\bys$ of the KITTI dataset was originally measured by a LiDAR scanner of Velodyne HDL-64E that collects sparse 3D points, and other unmeasured areas remain empty \citep{DBLP:conf/3dim/UhrigSSFBG17,DBLP:conf/iros/MaddernN16}. For the monocular depth estimation task, only a small portion of the pixels in $(n_h \times n_w)$ become valid pixels for use in training, whereas the other invalid pixels are ignored by applying a binary mask. Although the KITTI dataset provides a relatively denser ground-truth, other datasets, such as Argoverse or DDAD, exhibit substantially sparse ground-truth depth maps.

Second, the number of valid pixels is affected by the image crop. During the training of a monocular depth estimation model, several cropping methods, such as KB-crop, random-crop, and Garg-crop \citep{DBLP:conf/eccv/GargKC016}, are applied in a row to the image and depth map. Owing to the random-crop, the number of valid pixels contained in the ground-truth depth map varies every time. In a long training iteration, the cropped depth map can accidentally contain few valid pixels, such as $n=1$ or $n=0$, yielding a NaN loss when using an unbiased estimator of \texttt{torch.var(x)}. For sparse datasets such as the Argoverse or DDAD datasets, this phenomenon occurs in real situations (\tabref{tab:dataset}).

\begin{table}[t!]
	\centering
	\begin{tabular}{l|rr|r}
		\toprule
		Valid Pixels                                       & Avg.   & Min. & Valid Rate (\%) \\
		\midrule
		KITTI \citep{DBLP:journals/ijrr/GeigerLSU13}       & 25074  & 226  & 20.23           \\
		NYU-Depth V2 \citep{DBLP:conf/eccv/SilbermanHKF12} & 173598 & 6302 & 75.34           \\
		Driving Stereo \citep{DBLP:conf/cvpr/YangSHDSZ19}  & 31488  & 778  & 25.41           \\
		Argoverse \citep{DBLP:conf/cvpr/ChangLSSBHW0LRH19} & 1239   & 0    & 1.00            \\
		DDAD \citep{DBLP:conf/cvpr/GuiziliniAPRG20}        & 1489   & 0    & 1.20            \\
		\bottomrule
	\end{tabular}
	\caption{We measured the average and minimum number of valid pixels on cropped images generated during one epoch of training. We observed that the KITTI, NYU-Depth V2, and Driving Stereo datasets provide dense ground-truth depth maps. However, the Argoverse and DDAD datasets provide significantly sparse depth maps with a valid pixel rate of approximately 1\%, which can accidentally yield few valid pixels in the cropped image.}
	\label{tab:dataset}
\end{table}

\paragraph{Simulation} We simulated the impact of sparsity on the variance computation. First, we generate a random depth map of size $(n_h \times n_w)=(100 \times 100)$ and apply a binary mask to obtain a sparse depth map. Controlling its sparsity, we counted the number of NaNs when using $\texttt{torch.var(x)}$ and $\texttt{torch.var(x, unbiased=False)}$ on 10,000 simulations.

\begin{table}[t!]
	\centering
	\begin{tabular}{c|cc}
		\toprule
		                & \multicolumn{2}{c}{\# of NaNs for \texttt{torch.var(x)}}                           \\
		Valid Rate (\%) & Default                                                 & \texttt{unbiased=False} \\
		\midrule
		0.05            & 457                                                     & 92                      \\
		0.06            & 229                                                     & 30                      \\
		0.07            & 89                                                      & 8                       \\
		0.08            & 35                                                      & 5                       \\
		0.09            & 19                                                      & 1                       \\
		0.10            & 9                                                       & 1                       \\
		0.11            & 2                                                       & 0                       \\
		0.12            & 1                                                       & 0                       \\
		0.13            & 0                                                       & 0                       \\
		0.14            & 0                                                       & 0                       \\
		\bottomrule
	\end{tabular}
	\caption{Number of NaN occurrences with respect to the sparsity of valid pixels. The default option leads to an unbiased estimator and is vulnerable to $n=1$ and $n=0$, whereas specifying a biased estimator is only vulnerable to $n=0$, which can be passed in practice by checking the sanity of the depth map.}
	\label{tab:valid}
\end{table}

\tabref{tab:valid} summarizes the result. Generally, NaN is encountered only in cases of severe sparsity. We observed that when the rate of valid pixels was lower than 0.12\%, the variance computation began to yield NaN. Note that $\texttt{torch.var(x)}$ results in NaN for $n=1$ or $n=0$, whereas $\texttt{torch.var(x, unbiased=False)}$ yields NaN only for $n=0$, which can be passed in practice by verifying the sanity of the depth map. Because of this difference, $\texttt{torch.var(x)}$ is vulnerable to NaN loss. In consideration of this vulnerability, we claim the following:
\paragraph{Guideline 3.} We recommend using the mean-style implementation of scale-invariant log loss. When using the var-style implementation, do specify a biased estimator to obtain the correct results. Ensure skipping for $n=0$ by using a conditional statement.

\section{Experiments}
\label{sec:exp}

\begin{table*}[t!]
	\centering
	\resizebox{\textwidth}{!}{
		\begin{tabular}{lll|cccccccc}
			\toprule
			Dataset             & $\epsilon$                   & $\sigma_W$ & $\delta < 1.25 \uparrow$ & $\delta < 1.25^2 \uparrow$ & $\delta < 1.25^3 \uparrow$ & Abs Rel $\downarrow$ & Sq Rel $\downarrow$ & RMSE $\downarrow$ & RMSE log $\downarrow$ & log10 $\downarrow$ \\
			\midrule
			\multirow{13}{*}{K} & \multirow{13}{*}{0}          & 0.001$^*$  & 0.9751                   & 0.9974                     & 0.9994                     & 0.0514               & 0.1489              & 2.0690            & 0.0779                & 0.0224             \\
			                    &                              & 0.002      & 0.9755                   & 0.9974                     & 0.9994                     & 0.0514               & 0.1479              & 2.0628            & 0.0778                & 0.0225             \\
			                    &                              & 0.005      & 0.9754                   & 0.9974                     & 0.9993                     & 0.0513               & 0.1484              & 2.0644            & 0.0776                & 0.0223             \\
			                    &                              & 0.01       & 0.9755                   & \textbf{0.9975}            & 0.9994                     & 0.0510               & 0.1469              & 2.0572            & 0.0772                & 0.0223             \\
			                    &                              & 0.02       & 0.9757                   & 0.9974                     & 0.9994                     & 0.0511               & 0.1466              & 2.0549            & 0.0774                & 0.0223             \\
			                    &                              & 0.05       & 0.9755                   & 0.9974                     & 0.9994                     & 0.0514               & 0.1472              & 2.0475            & 0.0776                & 0.0224             \\
			                    &                              & 0.1        & \textbf{0.9761}          & 0.9973                     & 0.9994                     & \textbf{0.0508}      & \textbf{0.1454}     & \textbf{2.0226}   & \textbf{0.0767}       & \textbf{0.0221}    \\
			                    &                              & 0.2        & 0.9756                   & 0.9974                     & 0.9994                     & 0.0510               & 0.1467              & 2.0603            & 0.0774                & 0.0223             \\
			                    &                              & 0.5        & 0.9745                   & 0.9974                     & 0.9994                     & 0.0518               & 0.1494              & 2.0809            & 0.0783                & 0.0227             \\
			                    &                              & 1          & \multicolumn{8}{c}{NaN}                                                                                                                                                                          \\
			                    &                              & 2          & \multicolumn{8}{c}{NaN}                                                                                                                                                                          \\
			                    &                              & 5          & \multicolumn{8}{c}{NaN}                                                                                                                                                                          \\
			                    &                              & 10         & \multicolumn{8}{c}{NaN}                                                                                                                                                                          \\
			\midrule
			\multirow{13}{*}{K} & \multirow{13}{*}{$10^{-24}$} & 0.001      & 0.9756                   & 0.9974                     & 0.9994                     & 0.0520               & 0.1482              & 2.0616            & 0.0778                & 0.0226             \\
			                    &                              & 0.002      & 0.9756                   & 0.9974                     & 0.9994                     & 0.0514               & 0.1475              & 2.0385            & 0.0771                & \textbf{0.0222}    \\
			                    &                              & 0.005      & 0.9754                   & 0.9974                     & 0.9993                     & 0.0512               & 0.1474              & 2.0526            & 0.0776                & 0.0224             \\
			                    &                              & 0.01       & 0.9757                   & \textbf{0.9975}            & 0.9994                     & 0.0515               & 0.1463              & 2.0412            & 0.0772                & 0.0224             \\
			                    &                              & 0.02       & 0.9758                   & \textbf{0.9975}            & 0.9994                     & 0.0509               & 0.1459              & 2.0538            & \textbf{0.0770}       & \textbf{0.0222}    \\
			                    &                              & 0.05       & 0.9750                   & 0.9974                     & 0.9994                     & 0.0517               & 0.1480              & 2.0525            & 0.0778                & 0.0225             \\
			                    &                              & 0.1        & \textbf{0.9759}          & \textbf{0.9975}            & 0.9993                     & 0.0515               & 0.1469              & 2.0452            & 0.0775                & 0.0224             \\
			                    &                              & 0.2        & 0.9758                   & 0.9974                     & 0.9994                     & 0.0510               & 0.1469              & 2.0494            & 0.0773                & 0.0223             \\
			                    &                              & 0.5        & 0.9757                   & 0.9974                     & 0.9994                     & \textbf{0.0508}      & \textbf{0.1458}     & \textbf{2.0373}   & \textbf{0.0770}       & \textbf{0.0222}    \\
			                    &                              & 1          & 0.9751                   & 0.9974                     & 0.9994                     & 0.0515               & 0.1476              & 2.0583            & 0.0778                & 0.0225             \\
			                    &                              & 2          & 0.9736                   & 0.9971                     & 0.9993                     & 0.0545               & 0.1558              & 2.1027            & 0.0806                & 0.0238             \\
			                    &                              & 5          & 0.9721                   & 0.9968                     & 0.9992                     & 0.0570               & 0.1638              & 2.1590            & 0.0835                & 0.0250             \\
			                    &                              & 10         & 0.0070                   & 0.0198                     & 0.0402                     & 6.4862               & 453.7119            & 66.0504           & 1.9510                & 0.8128             \\
			\midrule
			\multirow{13}{*}{N} & \multirow{13}{*}{0}          & 0.001$^*$  & 0.9317                   & 0.9909                     & 0.9980                     & 0.0920               & 0.0449              & 0.3120            & 0.1124                & 0.0383             \\
			                    &                              & 0.002      & 0.9339                   & 0.9911                     & 0.9979                     & 0.0905               & 0.0460              & 0.3103            & 0.1113                & 0.0378             \\
			                    &                              & 0.005      & 0.9338                   & 0.9913                     & 0.9981                     & 0.0899               & 0.0443              & 0.3082            & 0.1110                & 0.0375             \\
			                    &                              & 0.01       & 0.9342                   & 0.9913                     & 0.9979                     & 0.0895               & 0.0451              & 0.3092            & 0.1107                & 0.0374             \\
			                    &                              & 0.02       & 0.9347                   & 0.9911                     & 0.9980                     & 0.0896               & 0.0447              & 0.3076            & 0.1107                & 0.0374             \\
			                    &                              & 0.05       & 0.9335                   & 0.9914                     & 0.9981                     & 0.0892               & 0.0447              & 0.3074            & 0.1104                & 0.0373             \\
			                    &                              & 0.1        & 0.9345                   & 0.9911                     & 0.9980                     & 0.0897               & 0.0448              & 0.3081            & 0.1108                & 0.0374             \\
			                    &                              & 0.2        & \textbf{0.9352}          & \textbf{0.9915}            & 0.9981                     & \textbf{0.0883}      & \textbf{0.0439}     & \textbf{0.3072}   & \textbf{0.1098}       & \textbf{0.0371}    \\
			                    &                              & 0.5        & \multicolumn{8}{c}{NaN}                                                                                                                                                                          \\
			                    &                              & 1          & \multicolumn{8}{c}{NaN}                                                                                                                                                                          \\
			                    &                              & 2          & \multicolumn{8}{c}{NaN}                                                                                                                                                                          \\
			                    &                              & 5          & \multicolumn{8}{c}{NaN}                                                                                                                                                                          \\
			                    &                              & 10         & \multicolumn{8}{c}{NaN}                                                                                                                                                                          \\
			\midrule
			\multirow{13}{*}{N} & \multirow{13}{*}{$10^{-24}$} & 0.001      & 0.9331                   & 0.9907                     & 0.9980                     & 0.0917               & 0.0460              & 0.3132            & 0.1123                & 0.0382             \\
			                    &                              & 0.002      & 0.9325                   & 0.9914                     & 0.9980                     & 0.0911               & 0.0452              & 0.3109            & 0.1119                & 0.0380             \\
			                    &                              & 0.005      & 0.9315                   & 0.9912                     & 0.9980                     & 0.0922               & 0.0456              & 0.3122            & 0.1126                & 0.0383             \\
			                    &                              & 0.01       & 0.9317                   & 0.9910                     & 0.9981                     & 0.0919               & 0.0451              & 0.3124            & 0.1126                & 0.0382             \\
			                    &                              & 0.02       & 0.9343                   & 0.9906                     & 0.9980                     & 0.0899               & 0.0446              & 0.3093            & 0.1111                & 0.0376             \\
			                    &                              & 0.05       & 0.9333                   & 0.9910                     & 0.9980                     & 0.0893               & 0.0443              & 0.3075            & 0.1107                & 0.0374             \\
			                    &                              & 0.1        & 0.9335                   & 0.9911                     & 0.9980                     & 0.0894               & 0.0450              & 0.3088            & 0.1107                & 0.0374             \\
			                    &                              & 0.2        & 0.9356                   & 0.9910                     & 0.9981                     & 0.0876               & \textbf{0.0430}     & \textbf{0.3045}   & 0.1094                & 0.0369             \\
			                    &                              & 0.5        & \textbf{0.9361}          & \textbf{0.9916}            & 0.9981                     & \textbf{0.0864}      & 0.0435              & 0.3046            & \textbf{0.1088}       & \textbf{0.0365}    \\
			                    &                              & 1          & 0.9340                   & 0.9910                     & 0.9980                     & 0.0888               & 0.0446              & 0.3073            & 0.1105                & 0.0372             \\
			                    &                              & 2          & 0.9296                   & 0.9909                     & 0.9981                     & 0.0907               & 0.0450              & 0.3130            & 0.1129                & 0.0381             \\
			                    &                              & 5          & 0.3129                   & 0.3424                     & 0.3679                     & 2.4012               & 18.8513             & 5.0322            & 1.0141                & 0.4224             \\
			                    &                              & 10         & 0.2678                   & 0.3328                     & 0.3660                     & 2.4188               & 18.8605             & 5.0928            & 1.0367                & 0.4304             \\
			\bottomrule
		\end{tabular}%
	}
	\caption{Experimental results of monocular depth estimation. Here, ``K'' denotes KITTI and ``N'' denotes NYU-Depth V2. $^*$ indicates the baseline.}
	\label{tab:exp}
\end{table*}

\paragraph{Objective} In the previous section, we demonstrated the occurrence of NaN loss in each of three vulnerabilities. Meanwhile, in the previous section, we claimed that allowing a high gradient scale enhances the performance of monocular depth estimation. To validate this claim, experiments were conducted with an extensive range of weight initializations. We pursued an investigation of the validity rather than hyperparameter tuning. The main objective of this study is to analyze and solve the problem of NaN loss. Achieving state-of-the-art performance is beyond the scope of this study.

\subsection{Implementation Details}

\paragraph{Target model} used in this study was the model studied by masked image modeling (MIM), which has recently achieved state-of-the-art performance in the monocular depth estimation task \citep{DBLP:conf/cvpr/XieGH00023}. The MIM successfully trained the SwinV2-Base \citep{DBLP:conf/cvpr/Liu0LYXWN000WG22} to achieve improved performance, which served as the pretrained backbone for the encoder of the monocular depth estimation network. We implemented the MIM using the official GitHub repository. Note that the official MIM implementation used weight initialization $\sigma_W=0.001$ without adding $\epsilon$; hence, we examined whether their weight initialization can be further improved in terms of stability and performance.

\paragraph{Hyperparameters} We used the AdamW optimizer \citep{DBLP:conf/iclr/LoshchilovH19} with $\beta_1=0.9$, $\beta_2=0.999$ and weight decay $5 \times 10^{-2}$, learning rate scheduler of polynomial decay using factor 0.9 and its maximum $5\times 10^{-4}$ to minimum $3\times 10^{-5}$, and number of epochs 25. The average of three runs with different random seeds is reported for each result. Training was conducted using a $4\times$A100 GPU machine.

\paragraph{Evaluation metrics} We used the following evaluation metrics commonly used in the monocular depth estimation tasks:
\begin{itemize}
	\item Threshold: \% of $y_i$ s.t. $\max(y_i/y_i^*, y_i^*/y_i)=\delta < thr$,
	\item Abs Rel: $\frac{1}{n} \sum_i{|y_i-y_i^*|/y_i^*}$,
	\item Sq Rel: $\frac{1}{n} \sum_i{(y_i-y_i^*)^2/y_i^*}$,
	\item RMSE: $\sqrt{\frac{1}{n} \sum_i{(y_i-y_i^*)^2}}$,
	\item RMSE log: $\sqrt{\frac{1}{n} \sum_i{(\log{y_i}-\log{y_i^*})^2}}$,
	\item log10: $\frac{1}{n} \sum_i{|\log_{10}{y_i}-\log_{10}{y_i^*}|}$.
\end{itemize}
Higher is better for the threshold metric, whereas lower is better for the other five metrics.

\paragraph{KITTI} dataset contains RGB images and the corresponding ground truth depth maps. The ground truth depth map was measured using a LiDAR scanner that collected sparse 3D points; the other unmeasured areas remained empty. In the KITTI dataset, the maximum depth was $M=80$. The KITTI dataset provides ground truth depth maps with an average valid pixel rate of approximately 20\% (\tabref{tab:dataset}). The size of the original image was $1241 \times 376$, which was cropped by KB-crop, random-crop, and Garg-crop during training to obtain a $352 \times 352$ image and a depth map.

\paragraph{NYU-Depth V2} dataset contains RGB images and their corresponding depth maps. The dataset was obtained from indoor scenes using a Microsoft Kinect, which provided dense ground truth depth maps with a valid pixel rate of approximately 75\%. For the NYU-Depth V2 dataset, the maximum depth was $M=10$. Following existing MIM practices, we used a $480 \times 480$ crop for training.

\subsection{Experimental Results} We set the last convolutional layer to $\nin=128$, $\nout=1$, and $k_h=k_w=3$. For Xavier initialization of $W \sim \N(0, 2/(k_h k_w (\nin + \nout)))$, we obtain $\sigma_W=0.0415$, while He initialization of $W \sim \N(0, 2/(k_h k_w \nout)))$ \citep{DBLP:conf/iccv/HeZRS15} yields $\sigma_W=0.4714$. However, it was observed that the best initialization could differ substantially from the two initializations, depending on the experimental setup.

We examined the effects of weight initialization. The first and third blocks of \tabref{tab:exp} summarize the results with $\epsilon=0$. We observed NaN loss when $\sigma_W \geq 1$ for KITTI. For NYU-Depth V2, it was observed that NaN loss occurred when using the initialization of $\sigma_W \geq 0.5$, which indicates that weight scale increased during training. Note that He initialization was close to this borderline. The best performance was observed at $\sigma_W=0.1$ for KITTI and $\sigma_W=0.2$ for NYU-Depth V2. Note that the latter is close to the NaN borderline of $\sigma_W=0.5$, and thus when using $\epsilon=0$, we should consider a trade-off between increasing $\sigma_W$ to obtain an active gradient and avoiding potential NaN loss.

We investigated the effect of adding $\epsilon=10^{-24}$. The second and fourth blocks of \tabref{tab:exp} summarize the results. Here no cases of NaN were observed, even in the previous NaN borderline of $\sigma_W=0.5$ or $\sigma_W=1$. Therefore, we can enjoy increasing $\sigma_W$ to obtain an active gradient without any potential NaN loss. However, as we described in the previous section, the stable initialization range changed by adding $\epsilon$. The best performance was obtained at a larger weight initialization of $\sigma_W = 0.5$ for KITTI and $\sigma_W=0.2$ or $\sigma_W=0.5$ for NYU-Depth V2.

\section{Conclusion}
\label{sec:con}
This study discussed three vulnerabilities in monocular depth estimation training. First, we found that the square root loss resulted in unstable gradient scaling. Vulnerability was found to cause NaN loss during the late training phase where we recommended using the original scale-invariant log loss. Second, we determined that the log-sigmoid function has numerical problems. The occurrence of NaN loss was demonstrated by module tests involving the log-sigmoid function, and two possible solutions were suggested. Finally, we revealed that half of the current implementations of the scale-invariant log loss yielded incorrect results owing to the use of an unbiased estimator. Because this problem can accidentally cause NaN loss in a sparse depth map, we claimed to use a biased estimator to ensure the correct result. Through experiments, we validated that the proposed guidelines improved optimization stability. We hope that our guidelines for obtaining stable optimization will aid the research community of monocular depth estimation.

{
    \small
    \bibliographystyle{ieeenat_fullname}
    \bibliography{main}

\begin{thebibliography}{51}
\providecommand{\natexlab}[1]{#1}
\providecommand{\url}[1]{\texttt{#1}}
\expandafter\ifx\csname urlstyle\endcsname\relax
  \providecommand{\doi}[1]{doi: #1}\else
  \providecommand{\doi}{doi: \begingroup \urlstyle{rm}\Url}\fi

\bibitem[Agarwal and Arora(2022)]{DBLP:conf/icip/Agarwal022}
Ashutosh Agarwal and Chetan Arora.
\newblock {Depthformer: Multiscale Vision Transformer for Monocular Depth
  Estimation with Global Local Information Fusion}.
\newblock In \emph{{ICIP}}, pages 3873--3877, 2022.

\bibitem[Agarwal and Arora(2023)]{DBLP:conf/wacv/AgarwalA23}
Ashutosh Agarwal and Chetan Arora.
\newblock {Attention Attention Everywhere: Monocular Depth Prediction with Skip
  Attention}.
\newblock In \emph{{WACV}}, pages 5850--5859, 2023.

\bibitem[Bhat et~al.(2021)Bhat, Alhashim, and Wonka]{DBLP:conf/cvpr/BhatAW21}
Shariq~Farooq Bhat, Ibraheem Alhashim, and Peter Wonka.
\newblock {AdaBins: Depth Estimation Using Adaptive Bins}.
\newblock In \emph{{CVPR}}, pages 4009--4018, 2021.

\bibitem[Bhat et~al.(2022)Bhat, Alhashim, and Wonka]{DBLP:conf/eccv/BhatAW22}
Shariq~Farooq Bhat, Ibraheem Alhashim, and Peter Wonka.
\newblock {LocalBins: Improving Depth Estimation by Learning Local
  Distributions}.
\newblock In \emph{{ECCV} {(1)}}, pages 480--496, 2022.

\bibitem[Bhat et~al.(2023)Bhat, Birkl, Wofk, Wonka, and
  M{\"{u}}ller]{DBLP:journals/corr/abs-2302-12288}
Shariq~Farooq Bhat, Reiner Birkl, Diana Wofk, Peter Wonka, and Matthias
  M{\"{u}}ller.
\newblock {ZoeDepth: Zero-shot Transfer by Combining Relative and Metric
  Depth}.
\newblock \emph{CoRR}, abs/2302.12288, 2023.

\bibitem[Brown(1947)]{brown1947small}
George~W Brown.
\newblock {On small-sample estimation}.
\newblock \emph{The Annals of Mathematical Statistics}, 18\penalty0
  (4):\penalty0 582--585, 1947.

\bibitem[Chang et~al.(2019)Chang, Lambert, Sangkloy, Singh, Bak, Hartnett,
  Wang, Carr, Lucey, Ramanan, and Hays]{DBLP:conf/cvpr/ChangLSSBHW0LRH19}
Ming{-}Fang Chang, John Lambert, Patsorn Sangkloy, Jagjeet Singh, Slawomir Bak,
  Andrew Hartnett, De Wang, Peter Carr, Simon Lucey, Deva Ramanan, and James
  Hays.
\newblock {Argoverse: 3D Tracking and Forecasting With Rich Maps}.
\newblock In \emph{{CVPR}}, pages 8748--8757, 2019.

\bibitem[Chen et~al.(2018)Chen, Zhu, Papandreou, Schroff, and
  Adam]{DBLP:conf/eccv/ChenZPSA18}
Liang{-}Chieh Chen, Yukun Zhu, George Papandreou, Florian Schroff, and Hartwig
  Adam.
\newblock {Encoder-Decoder with Atrous Separable Convolution for Semantic Image
  Segmentation}.
\newblock In \emph{{ECCV} {(7)}}, pages 833--851, 2018.

\bibitem[Dosovitskiy et~al.(2021)Dosovitskiy, Beyer, Kolesnikov, Weissenborn,
  Zhai, Unterthiner, Dehghani, Minderer, Heigold, Gelly, Uszkoreit, and
  Houlsby]{DBLP:conf/iclr/DosovitskiyB0WZ21}
Alexey Dosovitskiy, Lucas Beyer, Alexander Kolesnikov, Dirk Weissenborn,
  Xiaohua Zhai, Thomas Unterthiner, Mostafa Dehghani, Matthias Minderer, Georg
  Heigold, Sylvain Gelly, Jakob Uszkoreit, and Neil Houlsby.
\newblock {An Image is Worth 16x16 Words: Transformers for Image Recognition at
  Scale}.
\newblock In \emph{{ICLR}}, 2021.

\bibitem[Eigen et~al.(2014)Eigen, Puhrsch, and
  Fergus]{DBLP:conf/nips/EigenPF14}
David Eigen, Christian Puhrsch, and Rob Fergus.
\newblock {Depth Map Prediction from a Single Image using a Multi-Scale Deep
  Network}.
\newblock In \emph{{NIPS}}, pages 2366--2374, 2014.

\bibitem[Garg et~al.(2016)Garg, Kumar, Carneiro, and
  Reid]{DBLP:conf/eccv/GargKC016}
Ravi Garg, B.~G.~Vijay Kumar, Gustavo Carneiro, and Ian~D. Reid.
\newblock {Unsupervised {CNN} for Single View Depth Estimation: Geometry to the
  Rescue}.
\newblock In \emph{{ECCV} {(8)}}, pages 740--756, 2016.

\bibitem[Garipov et~al.(2018)Garipov, Izmailov, Podoprikhin, Vetrov, and
  Wilson]{DBLP:conf/nips/GaripovIPVW18}
Timur Garipov, Pavel Izmailov, Dmitrii Podoprikhin, Dmitry~P. Vetrov, and
  Andrew~Gordon Wilson.
\newblock {Loss Surfaces, Mode Connectivity, and Fast Ensembling of DNNs}.
\newblock In \emph{NeurIPS}, pages 8803--8812, 2018.

\bibitem[Geiger et~al.(2013)Geiger, Lenz, Stiller, and
  Urtasun]{DBLP:journals/ijrr/GeigerLSU13}
Andreas Geiger, Philip Lenz, Christoph Stiller, and Raquel Urtasun.
\newblock {Vision meets robotics: The {KITTI} dataset}.
\newblock \emph{Int. J. Robotics Res.}, 32\penalty0 (11):\penalty0 1231--1237,
  2013.

\bibitem[Glorot and Bengio(2010)]{DBLP:journals/jmlr/GlorotB10}
Xavier Glorot and Yoshua Bengio.
\newblock {Understanding the difficulty of training deep feedforward neural
  networks}.
\newblock In \emph{{AISTATS}}, pages 249--256, 2010.

\bibitem[Guizilini et~al.(2020)Guizilini, Ambrus, Pillai, Raventos, and
  Gaidon]{DBLP:conf/cvpr/GuiziliniAPRG20}
Vitor Guizilini, Rares Ambrus, Sudeep Pillai, Allan Raventos, and Adrien
  Gaidon.
\newblock {3D Packing for Self-Supervised Monocular Depth Estimation}.
\newblock In \emph{{CVPR}}, pages 2482--2491, 2020.

\bibitem[He et~al.(2015)He, Zhang, Ren, and Sun]{DBLP:conf/iccv/HeZRS15}
Kaiming He, Xiangyu Zhang, Shaoqing Ren, and Jian Sun.
\newblock {Delving Deep into Rectifiers: Surpassing Human-Level Performance on
  ImageNet Classification}.
\newblock In \emph{{ICCV}}, pages 1026--1034, 2015.

\bibitem[He et~al.(2016)He, Zhang, Ren, and Sun]{DBLP:conf/cvpr/HeZRS16}
Kaiming He, Xiangyu Zhang, Shaoqing Ren, and Jian Sun.
\newblock {Deep Residual Learning for Image Recognition}.
\newblock In \emph{{CVPR}}, pages 770--778, 2016.

\bibitem[Ioffe and Szegedy(2015)]{DBLP:conf/icml/IoffeS15}
Sergey Ioffe and Christian Szegedy.
\newblock {Batch Normalization: Accelerating Deep Network Training by Reducing
  Internal Covariate Shift}.
\newblock In \emph{{ICML}}, pages 448--456, 2015.

\bibitem[Ji et~al.(2023)Ji, Chen, Xie, Hong, Liu, Liu, Lu, Li, and
  Luo]{DBLP:journals/corr/abs-2303-17559}
Yuanfeng Ji, Zhe Chen, Enze Xie, Lanqing Hong, Xihui Liu, Zhaoqiang Liu, Tong
  Lu, Zhenguo Li, and Ping Luo.
\newblock {{DDP:} Diffusion Model for Dense Visual Prediction}.
\newblock \emph{CoRR}, abs/2303.17559, 2023.

\bibitem[Kim et~al.(2022)Kim, Ga, Ahn, Joo, Chun, and
  Kim]{DBLP:journals/corr/abs-2201-07436}
Doyeon Kim, Woonghyun Ga, Pyunghwan Ahn, Donggyu Joo, Sewhan Chun, and Junmo
  Kim.
\newblock {Global-Local Path Networks for Monocular Depth Estimation with
  Vertical CutDepth}.
\newblock \emph{CoRR}, abs/2201.07436, 2022.

\bibitem[Kingma and Ba(2015)]{DBLP:journals/corr/KingmaB14}
Diederik~P. Kingma and Jimmy Ba.
\newblock {Adam: {A} Method for Stochastic Optimization}.
\newblock In \emph{{ICLR}}, 2015.

\bibitem[Lee et~al.(2019)Lee, Han, Ko, and
  Suh]{DBLP:journals/corr/abs-1907-10326}
Jin~Han Lee, Myung{-}Kyu Han, Dong~Wook Ko, and Il~Hong Suh.
\newblock {From Big to Small: Multi-Scale Local Planar Guidance for Monocular
  Depth Estimation}.
\newblock \emph{CoRR}, abs/1907.10326, 2019.

\bibitem[Lewkowycz and Gur{-}Ari(2020)]{DBLP:conf/nips/LewkowyczG20}
Aitor Lewkowycz and Guy Gur{-}Ari.
\newblock {On the training dynamics of deep networks with L2 regularization}.
\newblock In \emph{NeurIPS}, 2020.

\bibitem[Li et~al.(2018)Li, Xu, Taylor, Studer, and
  Goldstein]{DBLP:conf/nips/Li0TSG18}
Hao Li, Zheng Xu, Gavin Taylor, Christoph Studer, and Tom Goldstein.
\newblock {Visualizing the Loss Landscape of Neural Nets}.
\newblock In \emph{NeurIPS}, pages 6391--6401, 2018.

\bibitem[Li et~al.(2023)Li, Chen, Liu, and
  Jiang]{DBLP:journals/corr/abs-2203-14211}
Zhenyu Li, Zehui Chen, Xianming Liu, and Junjun Jiang.
\newblock {DepthFormer: Exploiting Long-Range Correlation and Local Information
  for Accurate Monocular Depth Estimation}.
\newblock \emph{Machine Intelligence Research}, 2023.

\bibitem[Liu et~al.(2023)Liu, Kumar, Gu, Timofte, and
  Gool]{DBLP:conf/iclr/LiuKGTG23}
Ce Liu, Suryansh Kumar, Shuhang Gu, Radu Timofte, and Luc~Van Gool.
\newblock {VA-DepthNet: {A} Variational Approach to Single Image Depth
  Prediction}.
\newblock In \emph{{ICLR}}, 2023.

\bibitem[Liu et~al.(2022)Liu, Hu, Lin, Yao, Xie, Wei, Ning, Cao, Zhang, Dong,
  Wei, and Guo]{DBLP:conf/cvpr/Liu0LYXWN000WG22}
Ze Liu, Han Hu, Yutong Lin, Zhuliang Yao, Zhenda Xie, Yixuan Wei, Jia Ning, Yue
  Cao, Zheng Zhang, Li Dong, Furu Wei, and Baining Guo.
\newblock {Swin Transformer {V2:} Scaling Up Capacity and Resolution}.
\newblock In \emph{{CVPR}}, pages 11999--12009, 2022.

\bibitem[Loshchilov and Hutter(2019)]{DBLP:conf/iclr/LoshchilovH19}
Ilya Loshchilov and Frank Hutter.
\newblock {Decoupled Weight Decay Regularization}.
\newblock In \emph{{ICLR}}, 2019.

\bibitem[Maddern and Newman(2016)]{DBLP:conf/iros/MaddernN16}
Will Maddern and Paul~M. Newman.
\newblock {Real-time probabilistic fusion of sparse 3D {LIDAR} and dense
  stereo}.
\newblock In \emph{{IROS}}, pages 2181--2188, 2016.

\bibitem[Manimaran and Swaminathan(2022)]{9824488}
Gouthamaan Manimaran and J Swaminathan.
\newblock {Focal-WNet: An Architecture Unifying Convolution and Attention for
  Depth Estimation}.
\newblock In \emph{2022 IEEE 7th International conference for Convergence in
  Technology (I2CT)}, pages 1--7, 2022.

\bibitem[Ning et~al.(2023)Ning, Li, Zhang, Geng, Dai, He, and
  Hu]{DBLP:journals/corr/abs-2301-02229}
Jia Ning, Chen Li, Zheng Zhang, Zigang Geng, Qi Dai, Kun He, and Han Hu.
\newblock {All in Tokens: Unifying Output Space of Visual Tasks via Soft
  Token}.
\newblock \emph{CoRR}, abs/2301.02229, 2023.

\bibitem[Oquab et~al.(2023)Oquab, Darcet, Moutakanni, Vo, Szafraniec, Khalidov,
  Fernandez, Haziza, Massa, El{-}Nouby, Assran, Ballas, Galuba, Howes, Huang,
  Li, Misra, Rabbat, Sharma, Synnaeve, Xu, J{\'{e}}gou, Mairal, Labatut,
  Joulin, and Bojanowski]{DBLP:journals/corr/abs-2304-07193}
Maxime Oquab, Timoth{\'{e}}e Darcet, Th{\'{e}}o Moutakanni, Huy Vo, Marc
  Szafraniec, Vasil Khalidov, Pierre Fernandez, Daniel Haziza, Francisco Massa,
  Alaaeldin El{-}Nouby, Mahmoud Assran, Nicolas Ballas, Wojciech Galuba,
  Russell Howes, Po{-}Yao Huang, Shang{-}Wen Li, Ishan Misra, Michael~G.
  Rabbat, Vasu Sharma, Gabriel Synnaeve, Hu Xu, Herv{\'{e}} J{\'{e}}gou, Julien
  Mairal, Patrick Labatut, Armand Joulin, and Piotr Bojanowski.
\newblock {DINOv2: Learning Robust Visual Features without Supervision}.
\newblock \emph{CoRR}, abs/2304.07193, 2023.

\bibitem[Pascanu et~al.(2013)Pascanu, Mikolov, and
  Bengio]{DBLP:conf/icml/PascanuMB13}
Razvan Pascanu, Tom{\'{a}}s Mikolov, and Yoshua Bengio.
\newblock {On the difficulty of training recurrent neural networks}.
\newblock In \emph{{ICML} {(3)}}, pages 1310--1318, 2013.

\bibitem[Paszke et~al.(2019)Paszke, Gross, Massa, Lerer, Bradbury, Chanan,
  Killeen, Lin, Gimelshein, Antiga, Desmaison, K{\"{o}}pf, Yang, DeVito,
  Raison, Tejani, Chilamkurthy, Steiner, Fang, Bai, and
  Chintala]{DBLP:conf/nips/PaszkeGMLBCKLGA19}
Adam Paszke, Sam Gross, Francisco Massa, Adam Lerer, James Bradbury, Gregory
  Chanan, Trevor Killeen, Zeming Lin, Natalia Gimelshein, Luca Antiga, Alban
  Desmaison, Andreas K{\"{o}}pf, Edward~Z. Yang, Zachary DeVito, Martin Raison,
  Alykhan Tejani, Sasank Chilamkurthy, Benoit Steiner, Lu Fang, Junjie Bai, and
  Soumith Chintala.
\newblock {PyTorch: An Imperative Style, High-Performance Deep Learning
  Library}.
\newblock In \emph{NeurIPS}, pages 8024--8035, 2019.

\bibitem[Patil et~al.(2022)Patil, Sakaridis, Liniger, and
  Gool]{DBLP:conf/cvpr/PatilSLG22}
Vaishakh Patil, Christos Sakaridis, Alexander Liniger, and Luc~Van Gool.
\newblock {P3Depth: Monocular Depth Estimation with a Piecewise Planarity
  Prior}.
\newblock In \emph{{CVPR}}, pages 1600--1611, 2022.

\bibitem[Piccinelli et~al.(2023)Piccinelli, Sakaridis, and
  Yu]{DBLP:conf/cvpr/PiccinelliSY23}
Luigi Piccinelli, Christos Sakaridis, and Fisher Yu.
\newblock {iDisc: Internal Discretization for Monocular Depth Estimation}.
\newblock In \emph{{CVPR}}, pages 21477--21487, 2023.

\bibitem[Ranftl et~al.(2021)Ranftl, Bochkovskiy, and
  Koltun]{DBLP:conf/iccv/RanftlBK21}
Ren{\'{e}} Ranftl, Alexey Bochkovskiy, and Vladlen Koltun.
\newblock {Vision Transformers for Dense Prediction}.
\newblock In \emph{{ICCV}}, pages 12159--12168, 2021.

\bibitem[Reichmann(1961)]{reichmann1961use}
William~John Reichmann.
\newblock {Use and abuse of statistics}.
\newblock 1961.

\bibitem[Shao et~al.(2023{\natexlab{a}})Shao, Pei, Chen, Li, Liu, and
  Li]{DBLP:journals/corr/abs-2302-08149}
Shuwei Shao, Zhongcai Pei, Weihai Chen, Ran Li, Zhong Liu, and Zhengguo Li.
\newblock {URCDC-Depth: Uncertainty Rectified Cross-Distillation with CutFlip
  for Monocular Depth Estimation}.
\newblock \emph{CoRR}, abs/2302.08149, 2023{\natexlab{a}}.

\bibitem[Shao et~al.(2023{\natexlab{b}})Shao, Pei, Wu, Liu, Chen, and
  Li]{DBLP:journals/corr/abs-2309-14137}
Shuwei Shao, Zhongcai Pei, Xingming Wu, Zhong Liu, Weihai Chen, and Zhengguo
  Li.
\newblock {IEBins: Iterative Elastic Bins for Monocular Depth Estimation}.
\newblock In \emph{NeurIPS}, 2023{\natexlab{b}}.

\bibitem[Silberman et~al.(2012)Silberman, Hoiem, Kohli, and
  Fergus]{DBLP:conf/eccv/SilbermanHKF12}
Nathan Silberman, Derek Hoiem, Pushmeet Kohli, and Rob Fergus.
\newblock {Indoor Segmentation and Support Inference from {RGBD} Images}.
\newblock In \emph{{ECCV} {(5)}}, pages 746--760, 2012.

\bibitem[Song et~al.(2021)Song, Lim, and Kim]{DBLP:journals/tcsv/SongLK21}
Minsoo Song, Seokjae Lim, and Wonjun Kim.
\newblock {Monocular Depth Estimation Using Laplacian Pyramid-Based Depth
  Residuals}.
\newblock \emph{{IEEE} Trans. Circuits Syst. Video Technol.}, 31\penalty0
  (11):\penalty0 4381--4393, 2021.

\bibitem[Thompson et~al.(2021)Thompson, Phung, and
  Bouzerdoum]{DBLP:journals/access/ThompsonPB21}
Joshua~Luke Thompson, Son~Lam Phung, and Abdesselam Bouzerdoum.
\newblock {D-Net: {A} Generalised and Optimised Deep Network for Monocular
  Depth Estimation}.
\newblock \emph{{IEEE} Access}, 9:\penalty0 134543--134555, 2021.

\bibitem[Uhrig et~al.(2017)Uhrig, Schneider, Schneider, Franke, Brox, and
  Geiger]{DBLP:conf/3dim/UhrigSSFBG17}
Jonas Uhrig, Nick Schneider, Lukas Schneider, Uwe Franke, Thomas Brox, and
  Andreas Geiger.
\newblock {Sparsity Invariant CNNs}.
\newblock In \emph{3DV}, pages 11--20, 2017.

\bibitem[van Laarhoven(2017)]{DBLP:journals/corr/Laarhoven17b}
Twan van Laarhoven.
\newblock {{L2} Regularization versus Batch and Weight Normalization}.
\newblock \emph{CoRR}, abs/1706.05350, 2017.

\bibitem[Xie et~al.(2020)Xie, Sato, and
  Sugiyama]{DBLP:journals/corr/abs-2011-11152}
Zeke Xie, Issei Sato, and Masashi Sugiyama.
\newblock {Stable Weight Decay Regularization}.
\newblock \emph{CoRR}, abs/2011.11152, 2020.

\bibitem[Xie et~al.(2023)Xie, Geng, Hu, Zhang, Hu, and
  Cao]{DBLP:conf/cvpr/XieGH00023}
Zhenda Xie, Zigang Geng, Jingcheng Hu, Zheng Zhang, Han Hu, and Yue Cao.
\newblock {Revealing the Dark Secrets of Masked Image Modeling}.
\newblock In \emph{{CVPR}}, pages 14475--14485, 2023.

\bibitem[Yang et~al.(2019)Yang, Song, Huang, Deng, Shi, and
  Zhou]{DBLP:conf/cvpr/YangSHDSZ19}
Guorun Yang, Xiao Song, Chaoqin Huang, Zhidong Deng, Jianping Shi, and Bolei
  Zhou.
\newblock {DrivingStereo: {A} Large-Scale Dataset for Stereo Matching in
  Autonomous Driving Scenarios}.
\newblock In \emph{{CVPR}}, pages 899--908, 2019.

\bibitem[Yuan et~al.(2022)Yuan, Gu, Dai, Zhu, and
  Tan]{DBLP:conf/cvpr/Yuan0DZT22}
Weihao Yuan, Xiaodong Gu, Zuozhuo Dai, Siyu Zhu, and Ping Tan.
\newblock {Neural Window Fully-connected CRFs for Monocular Depth Estimation}.
\newblock In \emph{{CVPR}}, pages 3906--3915, 2022.

\bibitem[Zhang et~al.(2020)Zhang, He, Sra, and
  Jadbabaie]{DBLP:conf/iclr/ZhangHSJ20}
Jingzhao Zhang, Tianxing He, Suvrit Sra, and Ali Jadbabaie.
\newblock {Why Gradient Clipping Accelerates Training: {A} Theoretical
  Justification for Adaptivity}.
\newblock In \emph{{ICLR}}, 2020.

\bibitem[Zhao et~al.(2023)Zhao, Rao, Liu, Liu, Zhou, and
  Lu]{DBLP:journals/corr/abs-2303-02153}
Wenliang Zhao, Yongming Rao, Zuyan Liu, Benlin Liu, Jie Zhou, and Jiwen Lu.
\newblock {Unleashing Text-to-Image Diffusion Models for Visual Perception}.
\newblock In \emph{{ICCV}}, 2023.

\end{thebibliography}
}


\end{document}